# Exploring Dynamic Environments Using Stochastic Search Strategies

C.A. Piña-García[1], Dongbing Gu[2], J. Mario Siqueiros-Garca[1], Gustavo Carreón[1] and Carlos Gershenson[1,3,4,5,6]

[1]Instituto de Investigaciones en Matemáticas Aplicadas y en Sistemas, Universidad Nacional Autónoma de México, México, D.F.
[2]School of Computer Science and Electronic Engineering, University of Essex, Wivenhoe Park, Colchester, UK
[3]Centro de Ciencias de la Complejidad, Universidad Nacional Autónoma de México, México, D.F., México
[4]SENSEable City Lab, Massachusetts Institute of Technology Cambridge MA USA
[5]MoBS Lab, Network Science Institute, Northeastern University, Boston MA USA
[6]ITMO University, Birzhevaya liniya 4, St. Petersburg, Russian Federation
cgg@unam.mx

**Abstract**

In this paper, we conduct a literature review of laws of motion based on stochastic search strategies which are mainly focused on exploring highly dynamic environments. In this regard, stochastic search strategies represent an interesting alternative to cope with uncertainty and reduced perceptual capabilities. This study aims to present an introductory overview of research in terms of directional rules and searching methods mainly based on bio-inspired approaches.

This study critically examines the role of animal searching behavior applied to random walk models using stochastic rules and *kinesis* or *taxis*. The aim of this study is to examine existing techniques and to select relevant work on random walks and analyze their actual contributions. In this regard, we cover a wide range of displacement events with an orientation mechanism given by a reactive behavior or a source-seeking behavior. Finally, we conclude with a discussion concerning the usefulness of using optimal foraging strategies as a reliable methodology.

## Introduction

Stochastic search strategies plays an important role in terms of facing environmental uncertainty. Therefore, the present paper pretends to uncover the most insightful directional rules inspired by stochastic methods, statistical physics and random walks. Likewise, we consider these strategies as an emergent phenomenon (formation of global patterns from solely local interactions) which is a frequent and fascinating theme in the scientific literature both popular and academic (Downing, 2015).

The aim of this paper is to to examine existing techniques and do a comprehensive analysis to understand state-of-the-art, trends and research gaps. It is important to mention that these strategies can be used in the field of robotics as exploration and discovery algorithms with the aim to speeding up searching tasks.

Stochastic search strategies are mainly inspired by optimal foraging theory which involves animal search behavior as an alternative for facing highly dynamic environments. Thus, these strategies can be viewed as a correlated process which may consists of displacements only broken by successive reorientation events. Strategies such as: Lévy walk, ballistic motion and correlated random walk are well known examples of foraging strategies, which are subject to statistical properties derived from Lévy stochastic processes (Viswanathan et al., 1999; Bartumeus et al., 2005, 2008; Viswanathan et al., 2011).

A considerable amount of literature has been published on stochastic search strategies. These studies have been most extensively applied in the field of biology, particularly, in movement ecology (Morales et al., 2004; Viswanathan et al., 1999; Plank and Codling, 2009; Codling et al., 2010). However, an expressed interest has been shown by robotics researchers for adopting or mimicking specific behaviors belonging to optimal foraging theory (Viswanathan et al., 2011).

Recent evidence suggests that search strategies are mainly related to availability, quality and quantity of publicly accessible data on animal movement and Artificial Intelligence techniques. Recently, synthetic experiments have shown that what really matters is where the explorer diffuses, not the manner by which the explorer gets there (Viswanathan et al., 2011). We therefore, decided to concentrate on what we considered to be some of the more significant developments in stochastic search models. It is important to mention that for organization purposes, we have split in two categories as follows: stochastic rules and directional rules (*taxes*).

## Stochastic rules

In this section, we will explain a group of strategies related to biological foraging. In this regard, the traditional concept of random search plays an important role introduced by the optimal foraging theory (Viswanathan et al., 2011). Central to the entire discipline of optimal foraging is the concept of stochastic cause-effect response, which is determined by what we could call a "complex environment[1]" (Giuggioli and Bartumeus, 2010).

One of the most significant current hypothesis in the biological field states that natural mechanisms should drive foraging organisms to maximize their energy intake. This model is well known as *optimal foraging theory* (Viswanathan et al., 2011). Thus, a organism is either a searcher, e.g., forager, predator, parasite, pollinator or it is a target, e.g., prey or food. It is necessary here to clarify exactly what is meant by a searcher. In this study, the term searcher will be used as a computational or embodied agent. Throughout this paper the term uncorrelated refers to the direction of movement which is completely independent of the previous direction and unbiased refers to that there is no preferred direction, i.e., the direction moved at each step is completely random.

---
[1]This term refers to a highly dynamic environment, and it will be used indistinguishably in this paper.

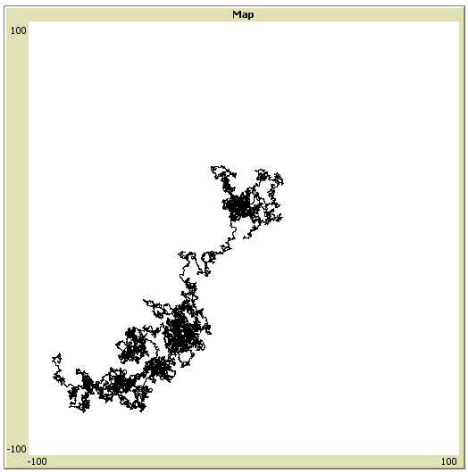

Figure 1: A two dimensional random walk showing a random walker using Brownian motion to explore a given region. The random walker chooses new regions to explore blindly and it has no any tendency to move toward regions that it has not occupied before.

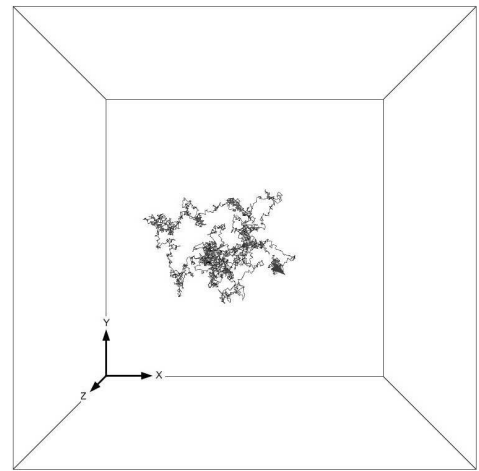

Figure 2: Plot of a Brownian motion in three dimensions after 15000 time steps.

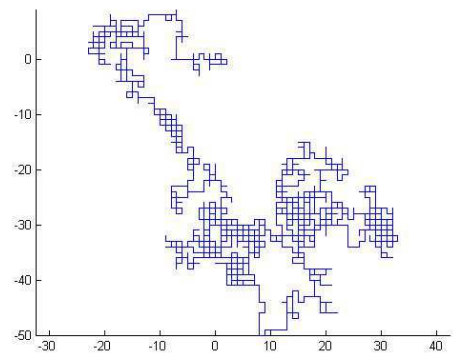

Figure 3: Plot of a two-dimensional lattice random walk. It requires all walking steps to exactly conform to the lattice of regular grid nodes and each move is restricted only to one of the adjacent nodes.

## Brownian motion as a random walk

The Brownian motion is one of the most widely used random walks and have been extensively applied for locating resources. Previous studies (Berg, 1993) have shown that this type of random walk presents explorations over short distances which can be made in much shorter times than explorations over long distances. In addition, the random walker tends to explore a given region of space and after that, it tends to return to the same point many times before finally wandering away. Therefore, it can be say that the random walker chooses new regions to explore blindly and it has no any tendency to move toward regions that it has not occupied before. In general, a Brownian walk has a normal diffusion where the mean square displacement increases in a linear way (Nurzaman et al., 1927). Fig. 1 presents a plot of simple Brownian walk.

An important point to note is that extensions to the Brownian model would include variable speed and/or turning frequency, waiting times between steps, temporally dependent parameters, interactions between individual walkers, and allowing movement in three dimensions rather than restricting the model to two dimensions (See Fig. 2).

## Lattice model as random walk

A lattice model can be considered as an arrangement of points or objects in a regular periodic pattern in two or three dimensions (Wang et al., 2009). The lattice space between successive steps of the random walk is constant, with every end point of the random walk being chosen as a grid node. It requires all walking steps to exactly conform to the lattice of regular grid nodes and each move is restricted only to one of the adjacent nodes (see Fig. 3 ).

A first serious discussion of lattice models can be described from the Chapman-Kolgomorov equations. This ap-proach can be applied to describe a random walk on a lattice (Viswanathan et al., 2011; Codling, 2003). Similarly, the Pauli master equation describing a random walker jumping between sites on a lattice is given by

$$\frac{d}{dt}P_k = \sum_{\ell}(W_{k,\ell}P_\ell - W_{\ell,k}P_k), \quad (1)$$

where $P_k(t)$ represents the probability of being in state $k$ at time $t$ and $W_{k,\ell}$ are the transition rates to go from site $\ell$ to site $k$. Thus, normal diffusion arises in the long time limit provided that (1) there is stochasticity (nondeterministic kinematics) and that (2) the transition rates have a range with finite variance (Viswanathan et al., 2011).

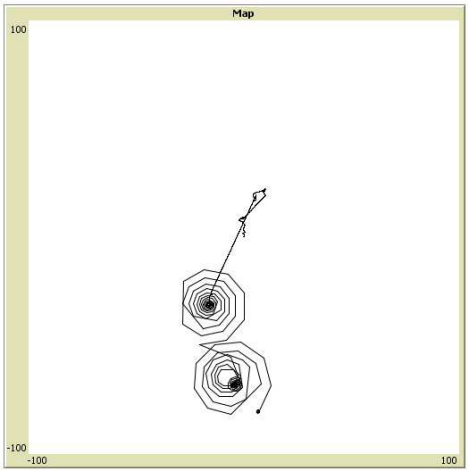

Figure 4: Plot of an Archimedean spiral showing a series of outwardly concentric circles connected by linear segments. A searcher returns repeatedly to the starting point of the search (homing behavior).

**Spiral searching**

There is a considerable amount of literature describing the role of systematic search rules that provide a set of alternative strategies to random walks (Reynolds et al., 2007). These studies point out that the classic example of a systematic search strategy is moving in the same direction. However, it is possible to find more complicated rules such as: the *Archimedean* spiral, also known as circular searches.

An important feature of this approach is the fact that under real conditions is very unlikely to obtain an almost perfect Archimedean spiral. A plot showing a spiral strategy is depicted in Fig. 4.

Similarly, according to (Bartumeus et al., 2005) the Archimedean spiral represents one of the most common searching rules in homing behaviors. In this case, a searcher is able to follow a spiral pattern during an initial phase, then systematically extends its range and moves in broad loops returning repeatedly to the starting point of the search as can be seen in Fig. 4. What is interesting in the results presented in (Zollner and Lima, 1999), is that spiral searching paths may work well in clumped landscapes.

In a controlled study of spiral searching, Reynolds in (Reynolds et al., 2007), reported that a spiral search can work in some cases where the navigation of the searcher were precise enough, and their visual detection ability were reliable enough to ensure that all areas are explored. However, in some cases targets could be missed without any chance of encountering them in a second round due to the path is an ever expanding spiral. Thus, relying on a spiral search pattern would be disastrous where navigational and detection systems are less than ideal. This method could be used for short searches, before the inevitable cumulative navigational error became too large, to allow a true spiral to be maintained (Press et al., 2007).

**Lévy walks**

There is a consensus among scientists that the search efficiency depends on the probability distribution of flight lengths taken by a searcher. These studies have reported that when the target sites are sparse, an inverse square power-law distribution of flight lengths corresponding to a Lévy flight, which can be considered as an optimal strategy.

Lévy walks are characterized by a distribution function $P(l_j) \sim l_j^{-\mu}$ with $1 < \mu \leq 3$ where $l_j$ is the flight length and the symbol $\sim$ refers to the asymptotic limiting behavior as the relevant quantity goes to infinity or zero (Viswanathan et al., 1999). For the special case when $\mu \geq 3$ a Gaussian or normal distribution arises due to the central limit theorem (Anderson et al., 1994).

The exponent of the power-law is named the Lévy index ($\mu$) and controls the range of correlations in the movement, introducing a family of distributions, ranging from Brownian motion ($\mu > 3$) to straight-line paths ($\mu \to 1$) (Bartumeus et al., 2005). In this regard, a random explorer can present an optimal strategy by selecting the Lévy index as follows: $\mu_{opt} = 2$. Thus, $\mu \approx 2$ is the optimal value for a search in any dimension. Fig. 5 provides an example of a Lévy walk using the probability distribution $P(l_j) \sim l_j^{-\mu}$ and $\mu \approx 2$.

An idealized model which captures some of the essential dynamics of foraging was developed by Viswanathan in (Viswanathan et al., 1999), where target sites are distributed randomly and the random walker behaves as follows:

1. If a target site lies within a "direct vision" distance $r_v$, then the random walker moves on a straight line to the nearest target site.

2. If there is no target site within a distance $r_v$, then the random walker chooses a direction at random and a distance $l_j$ from the probability distribution $P(l_j) \sim l_j^{-\mu}$. It then incrementally moves to the new point, constantly looking for a target within a radius $r_v$ along its way. If it does not detect a target, it stops after traversing the distance $l_j$ and chooses a new direction and a new distance $l_{j+1}$; otherwise, it proceeds to the target as in rule (1).

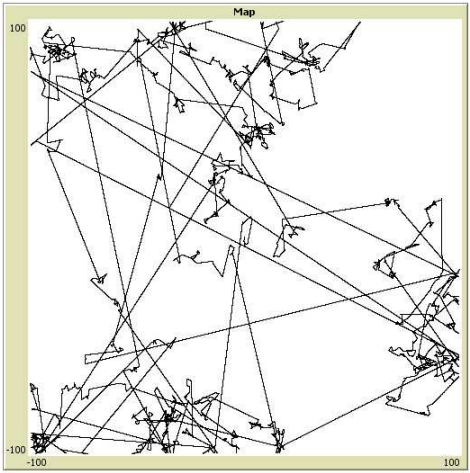

Figure 5: Plot of a Lévy walk where Lévy index: $\mu \approx 2$.

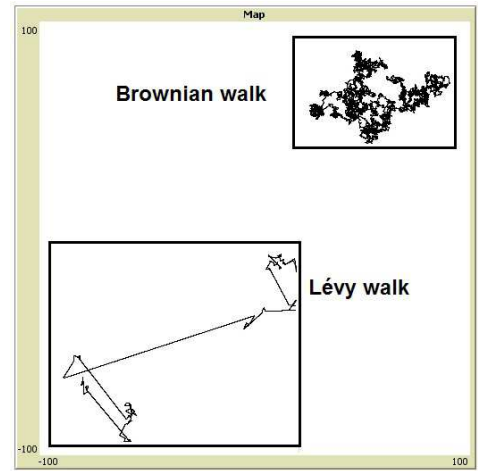

Figure 6: A Brownian walker returns many times to previously visited locations. In contrast, the Lévy walker frequently takes long jumps to unexplored territory.

Detailed examination of the mechanistic links between animal behavior, statistical patterns of movement and the role of randomness (stochasticity) in animal movement showed that temporal patterns can emerge at higher levels in terms of animal movement (Bartumeus, 2009, 2007). Thus, two important lessons can be learned from the model developed in (Viswanathan et al., 1999): (i) a search process of the Lévy-type may be optimal whenever the searcher have no information at all on the behavior of the target (e.g. location, velocity of movement, type of motion, etc.) even if targets are not uniformly distributed, and (ii)high directional persistence in the movement may not necessary be related to cues persecution.

According to Bartumeus in (Bartumeus et al., 2003) points out that the key advantage of a Lévy walk over other types of walks is restricted to prey density, mobility and size of the predator relative to the prey. In addition, a Lévy walk can be more efficient than the classic Brownian motion (Viswanathan et al., 1999, 2011). A comparison of the two walks reveals that Lévy walks do not consist simply in adding long walks to a Brownian motion i.e., these two types of motion differ in the whole flight-time probability distribution. Therefore, it is possible to claim that Lévy walks are better than Brownian walks when resources are scarce. However, it should be noted that Brownian motion is not necessarily a null model, it can be considered as a different searching strategy that is optimal under certain conditions. Fig. 6 compares two paths obtained from a Brownian motion and a Lévy walk.

**Correlated random walk**

The simplest way to incorporate directional persistence into a random walk model is introducing correlations (i.e., memory effects) between successive random walk steps (Viswanathan et al., 2011). Thus, the trajectories generated by correlated random walk models appear more similar to the empirical data than those generated by uncorrelated random walks. The correlated random walks (CRWs) appeared in the study of ecology when short and medium scaled animal movement was analyzed.

These CRWs have a correlation length or time that can be quantified via sinuosity and introduced into the random walk as follows: in two dimensions, typically two random walk step vectors differ only in their angular directions. The turning angles $\theta_j$ between successive step vectors $\mathbf{r}_j$ and $\mathbf{r}_{j+1}$ are usually chosen from a symmetric distribution. Thus, we can define the mean resultant as

$$\rho = \langle \cos(\theta) \rangle . \qquad (2)$$

It is important to note that CRW models have been studied in the context of biological mechanisms that link information processing directly to the sinuosity parameter $\rho$. In this regard, the correlation function for CRWs decay exponentially and hence the memory effects have a finite range. Consequently, beyond certain spatial and temporal scales CRWs become uncorrelated random walks and correlations are not strong enough, resulting in a loss of directional persistence at large spatial and temporal scales.

In another major study an explanatory theory which defines a correlated random walk is presented in (Bartumeus et al., 2005). This research argues that a CRW can be seen as a model that combines a Gaussian distribution of move lengths (i.e., displacement events) with a nonuniform angular distribution of turning angles (i.e., reorientation events). Likewise, it suggests that the optimization of random searches mainly depends on the optimal temporal execution of reorientation events.

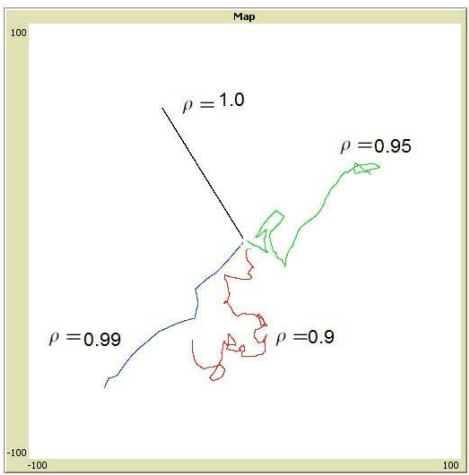

Figure 7: Plots showing various degrees of correlation generated by the shape parameter ($0 \leq \rho \leq 1$).

These CRW models are able to control directional persistence (i.e., the degree of correlation in the random walk) via the probability distribution of turning angles. Preliminary results using a wrapped Cauchy distribution (WCD) for the turning angles are reported in (Kent and Tyler, 1988). See equation 3.

$$\theta = \left[ 2 \times \arctan\left( \frac{(1-\rho) \times \tan(\pi \times (r-0.5))}{1+\rho} \right) \right]. \quad (3)$$

Where $\rho$ is the shape parameter ($0 \leq \rho \leq 1$) and $r$ is an uniformly distributed random variable $r \in [0,1]$. Directional persistence is controlled by changing the shape parameter of the WCD ($\rho$). Thus, for $\rho = 0$ we obtain an uniform distribution with no correlation between successive steps (Brownian motion), and for $\rho = 1$, we get a delta distribution at $0°$, leading to straight-line searches (see Fig. 7).

It is also important to note that the simplicity of random walks is methodologically attractive. However, these type of random searches result in redundant paths and may not be applicable to behaviorally sophisticated searchers. Therefore, many simulations tend to apply correlated random walks to reduce these redundancies and simulate more realistic movements.

## Directional rules

Recent developments in random walks have heightened the need for a better approach in terms of searching methods. In this regard, it has been suggested a suitable framework in terms of biological inspired motion e.g., birds, insects, mammals and parasites (Bartumeus and Catalan, 2009; Bartumeus, 2009; Bartumeus et al., 2005). Likewise, a set of important features need to be taken into account, for example: external factors (such as cues, obstacles and targets), information availability (full or partial) (Giuggioli and Bartumeus, 2010).

A large and growing body of literature has investigated a deterministic cause-effect strategies using an action-reaction behavior e.g., kinesis or taxis (Torney et al., 2009; Nicolau Jr et al., 2009; Russell et al., 2003; Tsuji et al., 2010).

In this section we will focus on examine a group of strategies that depend on attractants or repellents. Thus, random searchers tend to move according to certain chemicals attractants in their environment or towards the highest concentration of resources (biased movement and taxis). In addition, these rules can be successfully combined with stochastic rules with the aim to improve the performance of the searching task. It is important to note that these strategies are in principle based on a chemotactic phenomenon.

### Run-and-tumble chemotaxis

Chemotaxis can be defined as a biased random walk mainly composed of two phases. The "run phase" allows the cells to move with a constant velocity; and the "tumble phase" allows to reorient the cells to a new (random) direction (Viswanathan et al., 2011).

Recent evidence suggests that Monte Carlo simulations represents a suitable approach, for determining the effect of directional memory on the efficiency of "run-and-tumble" in environments with different attractant gradients (Nicolau Jr et al., 2009). Thus, to evaluate and compare results the authors designed a simple simulation as follows. First, $n$ bacteria were initially placed (and oriented) at random positions in a 2D simulation space. Then, attractant concentration was placed in this space with a linear gradient at point $(x, y)$, which is provided by

$$C(x,y) = max\left\{ C_{max} - k\sqrt{(x-x_0)^2 + (y-y_0)^2}, 0 \right\}. \quad (4)$$

Where $k$ is the gradient, $(x_0, y_0)$ is the origin and $C_{max}$ is the concentration of attractant at the peak of the gradient. In the presence of a gradient of attractant (or repellent), the bacteria use temporal comparisons of the attractant concentration over the preceding above 3–4 seconds to determine if conditions are improving or deteriorating. The cells compare their average receptor occupancy, approximated by the values of C, between 1 and 4 seconds ($s$) in the past, $\langle C \rangle_{1-4}$, to the average receptor occupancy during the past 1 s, $\langle C \rangle_{0-1}$, to produce the biaser $b = \langle C \rangle_{0-1} - \langle C \rangle_{1-4}$. If $b > 0$, the cell reduces the tumbling rate $\Gamma_{tumble}$ from the ambient value $\Gamma_0$ by an amount dependent on b: $\Gamma_{tumble} = \Gamma_0 - \gamma f(b)$, where $f(b)$ is a monotonically increasing function of b and $\gamma$ is a sensitivity coefficient that is positive for positive chemotaxis and negative for negative chemotaxis. The authors in (Nicolau Jr et al., 2009) set $\gamma$, the sensitivity of the response of the bacterium to changes in attractant concentration, to 1. Thus, If $b < 0$, $\Gamma_{tumble}$ is retained at the ambient value $\Gamma_0$. And if a bacterium $i$ is in a "run phase", its orientation $\theta_i$ and position $p_i$ are updated according to the system of equations

$$\theta_i(t+1) = \theta_i(t) + \eta D_{rot}, \quad (5)$$

$$p_i(t+1) = p_i(t) + v \begin{pmatrix} \cos\theta_i \\ \sin\theta_i \end{pmatrix}, \quad (6)$$

where $D_{rot}$ is the rotational diffusion coefficient (set to 0.15 $rad^2/s$), $\eta = \mathcal{N}(0,1)$ ($\mathcal{N}$ stands for normal distribution) and $v$ is the mean velocity of the bacterium. Consequently, $t$ is incremented by $\delta t$ (0.1s) and then, a random number $r$ between 0 and 1 is compared to $1/\Gamma_{tumble}$, if $r < 1/\Gamma_{tumble}$ then the cell tumbles and chooses a new direction

$$\theta_i(t+1) = \theta_i(t) + \vartheta_{tumble} D_{rot}, \quad (7)$$

where $\vartheta_{tumble}$ is the directional persistence parameter. If $\vartheta_{tumble} = 0$ then the new direction is identical to the previous direction (no reorientation). Large values of $\vartheta_{tumble}$ result in the new direction being independent of the old direction (perfect reorientation). Fig. 8 illustrates an example of the "run-and-tumble" chemotaxis strategy.

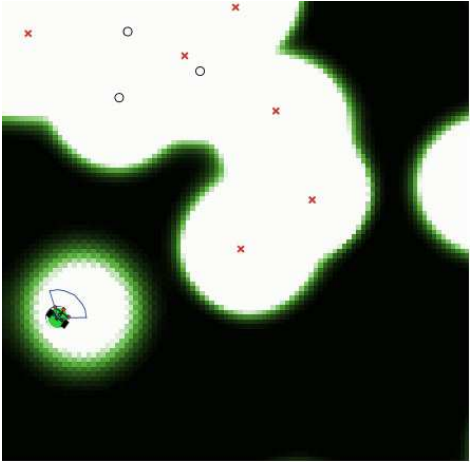

Figure 8: In this "run-and-tumble" model, a group of bacterial cells drift towards spatial regions with high nutrient concentration for growth and survival.

### Infotaxis

Infotaxis is focused on finding sources of particles transported in a random environment (Masson et al., 2009). The main issue is whether a source of particles can be located when the only clues of its presence are rare detections?. An good attempt to answer the question is a study carried out by Vergassola in (Vergassola et al., 2007).

It is possible to define the infotaxis strategy as a searching algorithm designed to work under an environment with sporadic cues and partial information; the process can be thought of as acquisition of information on source location, so that information plays an important role similar to concentration in chemotaxis. Consequently, the infotaxis strategy locally maximizes the expected rate of information gain. In general terms, the infotaxis can be considered as a strategy for searching without gradients. Thus, two main aspects must be taken into account. On one hand, new actions should be tried and the available phase space must be explored ( "exploration"). On the other hand, this should not be done blindly, e.g., the searcher in this case needs to minimize the searching time ( "exploitation"). Therefore, infotaxis can be applied more broadly in the context of searching with sparse information.

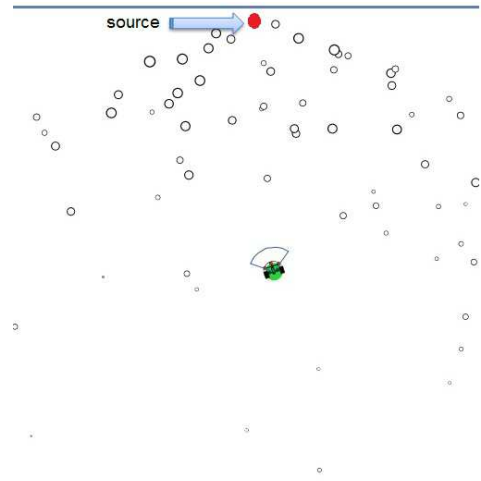

Figure 9: A computer simulation showing a searcher pausing to gather more information. Thus, it can obtain a more reliable estimate of the source distribution.

In (Vergassola et al., 2007) is argued that given a probability distribution $P(\mathbf{r}_0)$ for the location of the source, it is possible to show that the expected search time $\langle T \rangle$ is bounded by $\langle T \rangle \geq e^{S-1}$, where $S$ is the entropy of Shannon for the distribution $S \equiv -\int dx P(x) \ln P(x)$. The latter quantifies how spread-out the distribution is and goes to zero when the position of the source is localized to one site that is known. The rate of acquisition of information is quantified by the rate of reduction of entropy. Consequently, the main problem for the searcher is that the real probability distribution is unknown (to it) and must be estimated from the available data. As information accumulates, the entropy of the estimated distribution decreases and with it the expected time to locate the source. The searcher is faced with conflicting choices of either proceeding with its current information, or alternatively pausing to gather more information and obtain a more reliable estimate of the source distribution.

An important point to note is that in the search context, "exploitation" of the currently estimated $P_t(\mathbf{r}_0)$ by chasing locations of maximal estimated probability is very risky, because it can lead off the track. On the other hand, the most conservative "exploration" approach is to accumulate information before taking any step. This strategy is safe but not productive and is inferior to more active exploration, for example, systematic search in a particular sector. Hence, to balance exploration and exploitation, the searching algorithm needs to be redefined as follows: at each time step, the searcher chooses the direction that locally maximizes the expected rate of information acquisition. Specifically, the searcher chooses among the neighboring sites on a lattice and standing still, the move that maximizes the expected reduction in entropy of the posterior probability field. The intuitive idea is that entropy decreases (and thus information accumulates) faster close to the source because cues arrive at a higher rate. Consequently, tracking the maximum rate of information acquisition will guide the searcher to the source (see Fig. 9).

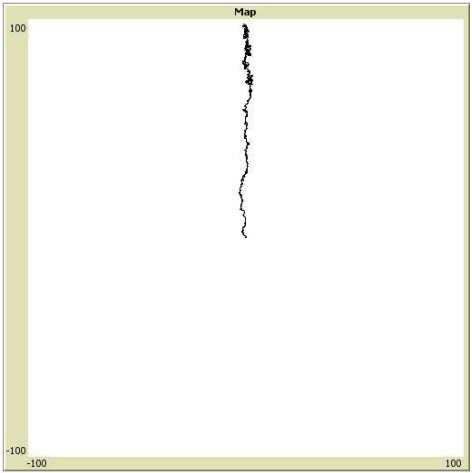

Figure 10: A plot showing an infotactic trajectory toward an emitting source. When the searcher is close to the source it starts to wander. The starting point was set to (0,0).

The authors in (Vergassola et al., 2007), estimate the variation of entropy expected upon moving to one of the neighboring points $\mathbf{r}_j$ (or standing still) as:

$$\overline{\Delta S}(\mathbf{r} \mapsto \mathbf{r}_j) = P_t(\mathbf{r}_j)\,[-S] + [1 - P_t(\mathbf{r}_j)][\rho_0(\mathbf{r}_j)\Delta S_0 + \rho_1(\mathbf{r}_j)\Delta S_1 + \ldots] \quad (8)$$

The first term on the right-hand side corresponds to finding the source, that is, $P_{t+1}$ becoming a $\delta$ function and entropy becoming zero. The second term on the right-hand side refers to the alternative case when the source is not found at $\mathbf{r}_j$. Symbols $\rho_k(\mathbf{r}_j)$ denote the probability that $k$ detections be made at $\mathbf{r}_j$ during a time-step $\Delta t$. The symbols $\Delta S_k$ in the equation denote the change of entropy between the fields $P_{t+1}(\mathbf{r}_0)$ and $P_t(\mathbf{r}_0)$. The first term on the right-hand side of the equation is the exploitative term, weighing only the event that the source is found at the point $\mathbf{r}_j$ and favoring motion to maximum likelihood points. The second term on the right-hand side of the equation is the information gain from receiving additional cues. Thus, infotaxis naturally combines exploitative and exploratory tendencies (Vergassola et al., 2007).

As discussed previously, the gain mostly arises when the searcher is close to the source and its wandering is slightly reduced. Fig. 10 provides an infotactic trajectory toward an emitting source.

An important point to note is that having multiple sources can generate conflicts and ambiguities. In this case, the searcher may get stuck by contradictory clues making its task more difficult (Masson et al., 2009).

## Discussion

The main problem encountered with all these strategies is that it is not possible to find a general methodology to cope with different environments, so that a searcher is highly sensitive to its sensing and orientating abilities. In this regard, research in the area of random walks is far from complete (Codling, 2003). There remains a wealth of exploration problems relating to random walks and searching strategies that have yet to be solved (for example, how to select the best strategy for exploring a complex environment) and, of course, an almost endless supply of biological systems that are amenable to modeling using exploration techniques.

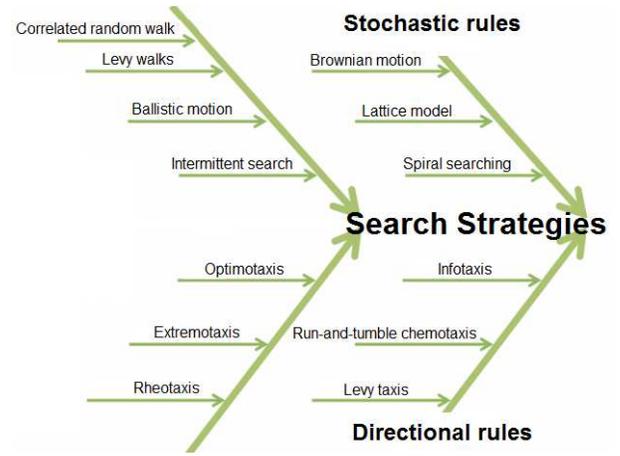

Figure 11: A diagram showing search strategies divided into two categories: stochastic rules and directional rules.

Similarly, according to the No Free Lunch Theorem of Optimization (NFLT) (Wolpert and Macready, 1997), which states that there is no one model that works best for every problem. The assumptions of a great model for one problem may not hold for another problem, so it is common in this searching context to try multiple strategies or models and find one that works best for a particular problem or environment.

Fig. 11 presents an overview of some existing techniques based on stochastic search strategies. These strategies are divided into two categories: stochastic rules and directional rules. It is important to note that we have added a set of extra *taxis* rules such as: Lévy taxis (Pasternak et al., 2009), Optimotaxis (Mesquita et al., 2008), Extremotaxis (Burrage et al., 2008) and Rheotaxis (Carton and Montgomery, 2003).

## Conclusions

In this paper, we have conducted an introductory overview of research of some existing techniques related to laws of motion based on stochastic search strategies. We have classified this study in two main categories: stochastic rules and directional rules, similarly, this survey has presented a variety of computer algorithms that have been assessed in terms of simulation. This manuscript has explained the central importance of using bio-inspired strategies with the aim to enhance our understanding of stochastic search strategies.

This work contributes to existing knowledge by adding and updating with current literature that provide with several practical applications. A number of caveats need to be noted regarding the present study. For instance, the random searcher is highly sensitive to its sensing and orientating abilities. Small changes in the environment can have a critical impact on the probability of success. In such case, the variability in the environment cannot be ignored.

Finally, We are well aware that there are many other techniques that we have not mentioned in this work. However, we consider that this introductory review is a good start point. Please refer to (García, 2014) for an in depth analysis about searching strategies.